\documentclass[conference, final]{IEEEtran}
\IEEEoverridecommandlockouts
\usepackage{cite}
\usepackage{amsmath,amssymb,amsfonts}
\usepackage{algorithmic}
\usepackage{graphicx}
\usepackage{textcomp}
\usepackage{xcolor}

\usepackage{arydshln}
\usepackage{dsfont}
\usepackage{hyperref} 
\usepackage{bbm}
\usepackage{url}            
\usepackage{amsfonts}       
\usepackage{nicefrac}       
\usepackage{microtype}      
\usepackage{graphicx}
\usepackage{subfigure}
\usepackage{setspace}
\usepackage{multirow}
\usepackage{cite}
\usepackage{arydshln}
\usepackage{dsfont}
\usepackage{arydshln}
\usepackage{wrapfig,lipsum,booktabs}
\usepackage{tcolorbox}
\usepackage{mdframed}
\usepackage{amsthm}

\usepackage{pgfplots}

\newtcolorbox{thm}{boxrule=0.0pt, arc=0pt, colback=grey}
\global\mdfdefinestyle{thm}{%
linecolor=black, backgroundcolor=gray!20, linewidth=0.0pt}

\newtheorem{theorem}{Theorem}

\DeclareMathOperator*{\argmax}{arg\,max}

\usepackage{array}

\newcolumntype{P}[1]{>{\centering\arraybackslash}p{#1}}

\newcommand \Reals {{\mathds{R}}}

\def\BibTeX{{\rm B\kern-.05em{\sc i\kern-.025em b}\kern-.08em
    T\kern-.1667em\lower.7ex\hbox{E}\kern-.125emX}}
\begin{document}

\title{Maximizing Influence with Graph Neural Networks
\thanks{Supported in part by ANR (French National Research Agency) under the JCJC project GraphIA (ANR-20-CE23-0009-01).}
}

\author{\IEEEauthorblockN{George Panagopoulos$^1$, Nikolaos Tziortziotis$^2$, Michalis Vazirgiannis$^1$, Fragkiskos D. Malliaros$^3$}
\IEEEauthorblockA{\textit{$^1$École Polytechnique, IP Paris, France}\\
\{george.panagopoulos, michalis.vazirgiannis\}@polytechnique.edu \\
\textit{$^2$Jellyfish, France}\\
ntziorzi@gmail.com \\
\textit{$^3$Université Paris-Saclay, CentraleSupélec, Inria, France}\\
fragkiskos.malliaros@centralesupelec.fr \\
}
}
\maketitle

\begin{abstract}
Finding the seed set that maximizes the influence spread over a network is a well-known NP-hard problem.
Though a greedy algorithm can provide near-optimal solutions, the subproblem of influence estimation renders the solutions inefficient.
In this work, we propose \textsc{Glie}, a graph neural network that learns how to estimate the influence spread of the independent cascade. 
\textsc{Glie} relies on a theoretical upper bound that is tightened through supervised training.
Experiments indicate that it provides accurate influence estimation for real graphs up to 10 times larger than the train set.
Subsequently, we incorporate it into two influence maximization techniques.
We first utilize Cost Effective Lazy Forward optimization substituting Monte Carlo simulations with \textsc{Glie}, surpassing the benchmarks albeit with a computational overhead.
To improve computational efficiency we develop a provably submodular influence spread based on \textsc{Glie}'s representations, 
to rank nodes while building the seed set adaptively. 
The proposed algorithms are inductive, meaning they are trained on graphs with less than 300 nodes and up to 5 seeds, 
and tested on graphs with millions of nodes and up to 200 seeds.
The final method exhibits the most promising combination of time efficiency and influence quality, outperforming several baselines.
\end{abstract}


\section{Introduction}
Several real-world problems can be cast as combinatorial optimization problem over a graph.
From distributing packages \cite{mathew2015planning} and vehicles' management \cite{touati2012combinatorial} optimization on graphs lies in the core of many real-world problems that are vital to our way of living.
Unfortunately, the majority of these problems are NP-hard, and hence we can only approximate their solution in a satisfactory time limit that matches the real world requirements. 
Recent machine learning methods have emerged as a promising solution to develop heuristic methods that provide fast and accurate approximations.
The general idea is to train a supervised or unsupervised learning model to infer the solution given an unseen graph and the problem constraints. The models tend to consist of Graph Neural Networks (GNNs) to encode the graph and the nodes, Q-learning \cite{sutton2018reinforcement} to produce sequential predictions, or a combination of both.
The practical motivation behind learning to solve combinatorial optimization problems, is that inference time is faster than running an exact combinatorial solver \cite{joshi2019learning}.
That said, specialized combinatorial algorithms like \textsc{Concorde} for the \textit{Traveling Salesman  Problem} (TSP) or \textsc{Gurobi} in general, cannot be surpassed yet \cite{kool2018attention}.

Though many such methods have been proposed for a plethora of problems, influence maximization (IM) has not been addressed yet extensively.
IM addresses the problem of finding the set of nodes in a network that would maximize the number of nodes reached by starting a diffusion from them \cite{kempe2003maximizing}. The problem is proved to be NP-hard, from a reduction to the set-cover problem. Moreover, the influence estimation (IE) problem that is embedded in IM, i.e., estimating the number of nodes influenced by a given seed set, is \#P-hard 
and would require $2^{|E|}$ possible combinations to compute exactly, where $|E|$ is the number of network edges \cite{wang2012scalable}. Typically, influence estimation is approximated using repetitive Monte-Carlo (MC) simulations of the independent cascade (IC) diffusion model \cite{tian2022unifying}. In general, the seed set is built greedily, taking advantage of the submodularity of the influence function which guarantees at least $(1-\frac{1}{e})$ approximation to the optimal. Although the latter lack of efficiency as one still has to estimate the influence of every candidate seed in every step of building the seed sets. Hence, several scalable algorithms \cite{borgs2014maximizing,tang2015influence} and heuristics \cite{chen2009efficient} were developed capitalizing on sketches or the structure of the graph to produce more efficient solutions.

We address IM using graph neural networks to capitalize on the aforementioned advantages and their ability to easily incorporate contextual information such as user profiles and topics \cite{tian2020deep}, a task that remains unsolvable for non-specialized IM algorithms and heuristics. We propose \textsc{Glie}, a GNN that provides efficient IE for a given seed set and a graph with influence probabilities. It can be used as a standalone influence predictor with competitive results for graphs up to 10 times larger than the train set. Moreover, we leverage \textsc{Glie} for IM, combining it with \textsc{Celf} \cite{leskovec2007cost},  that typically does not scale beyond networks with thousands of edges. The proposed method runs in networks with millions of edges in seconds, and exhibits better influence spread than a state-of-the-art algorithm and previous GNN-RL methods for IM. 
In addition, we propose \textsc{Pun}, a method that uses \textsc{Glie}'s representations to compute the number of neighbors predicted to be uninfluenced and uses it as an approximation to the marginal gain. We prove \textsc{Pun}'s influence spread is submodular and monotone, and hence can be optimized greedily with a guarantee, in contrast to prior learning-based methods. The experiments indicate that \textsc{Pun} provides the best balance between influence quality and  efficiency. 

The paper is organized as follows. Section~\ref{sec:related_work} presents an overview of relevant approaches and clarifies the proposed models' advantages. 
Section~\ref{sec:methodology} describes the proposed methods, starting with IE and advancing progressively towards faster methods for IM.
Section~\ref{sec:experiments} presents the experimental results for IE and IM. 
Finally, Section~\ref{sec:conclusion} summarizes the contribution and presents future steps.

\section{Related Work}
\label{sec:related_work}
The first approach to solving combinatorial optimization (CO) using neural networks was based on attention NNs for discrete structures, \textsc{PointerNets} \cite{vinyals2015pointer}, followed by an architecture that combines \textsc{PointerNets} with an actor-critic training to find the best route for TSP \cite{bello2016neural}.
The first architecture that utilized graph-based learning was \textsc{S2n-Dqn} \cite{dai2017learning}, using \textsc{Struct2Vec} to encode the states of the nodes and the graph, and training a Deep Q-network (\textsc{Dqn}) model that chooses the right node to add in a solution given the current state.

Based on \textsc{S2v-Dqn}, a \textsc{Dqn} 
for the network dismantling problem was recently proposed \cite{fan2020finding}. The model, named \textsc{Finder}, uses a deep Q-learning architecture where the representations are derived by three \textsc{GraphSage} layers. The reward is based on the size of the giant connected component size, i.e., every new node (seed) chosen, aims to dismantle the network as much as possible. 
Some of the main advantages of \textsc{Finder} is that it is trained on small synthetic data, which are easy to make, and can extrapolate to relatively large graphs.
On the other hand, one of the core disadvantages is that it can not work with directed graphs and weighted edges. 
Another recent supervised deep learning approach on IM, \textsc{Gcomb} \cite{manchanda2020gcomb}, utilizes a probabilistic greedy to produce scores on graphs and trains a GNN to predict them. A Q-network receives the scores along with an approximate calculation of the node's neighborhood correlation with the seed set,  to predict the next seed. 
This approach, though scalable and comparable to SOTA in accuracy, has to be trained on a large random subset of the graph ($30\%$ of it) and tested on the rest.
This makes the model graph-specific, i.e., it has to be retrained to perform well on a new graph. This imposes a serious overhead, considering the time required for training, subsampling and labeling these samples using the probabilistic greedy method with traditional IE. As shown in \cite{manchanda2020gcomb} Appendix G, it takes more than hundreds of minutes and is thus out of our scope.
Another GNN that addresses influence prediction is \cite{xia2021deepis}. \textsc{DeepIS} uses the power sequence of the influence probability matrix and a two-layer GNN to regress the susceptibility of each node. Subsequently, the estimation propagates in the neighbors based on the IC probability. \textsc{DeepIS} is a different architecture then \textsc{Glie}, which receives only indications of the seed set. Moreover, \textsc{DeepIS} is not tested extensively in influence maximization and as we will see in the experimental section, its use of the powers of influence probability matrix is detrimental to its scalability.
Finally, a recent work 
on learning approximations to general submodular policies \cite{alieva2020learning} requires a specific model to capture the state of IM, which is non-trivial to devise.
A different branch of learning-based IM relies on supplementary information such as diffusion cascades\cite{panagopoulos2018diffugreedy} to derive a more effective IM algorithm \cite{panagopoulos2020influence}. This is clearly diagonal to the current methodology, which does not assume any further information from the typical IM setting.

In this paper, we propose an approach that combines the advantages of the aforementioned methods in that it is only trained on small simulated data once and generalizes to larger graphs, and it addresses the problem of IM in weighted directed networks. Furthermore, the approach can be broken down into a GNN for influence estimation and two IM methods. The former can act alone as an influence predictor and be competitive with relevant methods, such as \textsc{Dmp} \cite{lokhov2019scalable} for graphs up to one scale larger than the train set. \textsc{Glie} is used to propose: (1) \textsc{Celf-Glie}, \textsc{Celf} \cite{leskovec2007cost} with \textsc{Glie} as influence estimator; (2)  \textsc{Pun}, an adaptive IM method \cite{AIMtutorialCautis19} that optimizes greedily a submodular influence spread using \textsc{Glie}'s representations.   

We note here that the majority of the relevant literature on deep learning for combinatorial optimization address small graphs \cite{vinyals2015pointer,dai2017learning} which makes them not applicable to our task.  More scalable, unsupervised methods \cite{karalias2020erdos} are tailored to specific problems and are non-trivial to adjust to our problem, with the exception of \cite{li2018combinatorial} which was found significantly worse than the SOTA algorithm we compare with\cite{manchanda2020gcomb}.

\section{Methodology}
\label{sec:methodology}

\subsection{{\normalfont{\textsc{Glie}}}: Graph Learning-based Influence Estimation}
\label{sec:glie}
In this section, we introduce \textsc{Glie}, a GNN model that aims to learn how to estimate the influence of seed set $S$ over a graph $G=(V,E)$.
Let $A \in \Reals^{n \times n}$ be the adjacency matrix and ${X} \in \Reals^{n \times d}$ be the features of nodes, representing which nodes belong to the seed set by $1$ and $0$ otherwise:
\begin{equation}
    X_u = \left\{ 
\begin{array}{cr}
\{1\}^d,& u \in S\\
      \{0\}^d,& u \notin S\\
\end{array}\right..
\label{eq:xdef}
\end{equation}

\noindent For the analysis that follows, we set $d=1$. More dimensions will become meaningful when we parameterize the problem.
If we normalize $A$ by each row, we form a row-stochastic transition matrix, as: 
\begin{equation}
A_{uv} = p_{vu} = \left\{ 
\begin{array}{cr}
\frac{1}{\text{deg}(u)}, & v \in \mathcal{N}(u) \\
      0, & v \notin \mathcal{N}(u) \\
\end{array}\right.,
\end{equation}
where $\text{deg}(u)$ is the in-degree of node $u$ and $\mathcal{N}(u) $ is the set of neighbors  of $u$. Based on the weighted cascade model \cite{kempe2003maximizing}, each row $u$ stores the probability of node $u$ being influenced by each of the other nodes that are connected to it by a directed link $v\rightarrow u$. Note that, in case of directed influence graphs, $A$ should correspond to the \textit{transpose} of the adjacency matrix. The influence probability $p(u|S)$ resembles the probability of a node $u$ getting influenced if its neighbors belong in the seed set, i.e., during the first step of the diffusion.
We can use message passing to compute a well-known upper bound $\hat{p}(u|S)$ of $p(u|S)$ for  $u$:
\begin{align}
     \hat{p}(u|S) = A_u \cdot  X & = \sum_{v \in \mathcal{N}(u)  \cap S} \frac{1}{\text{deg}(u)} = \\
    \sum_{v \in \mathcal{N}(u)  \cap S} p_{vu} & \geq 
    1- \prod_{v \in \mathcal{N}(u)  \cap S} (1-p_{vu}) = p(u|S),
    \label{eq:bound1}
\end{align}
where the second equality stems from the definition of the weighted cascade and the inequality from the proof in \cite{zhou2015upper}, App. A.
As the diffusion covers more than one-hop, the derivation requires repeating the multiplication to approximate the total influence spread. To be specific, computing the influence probability of nodes that are not adjacent to the seed set requires estimating recursively the probability of their neighbors being influenced by the seeds. If we let $H_1 = A \cdot  X$, and we assume the new seed set $S^t$ to be the nodes influenced in the step $t-1$, their probabilities are stored in $H_t$, much like a diffusion in discrete time. We can then recompute the new influence probabilities with $H_{t+1} = A \cdot  H_t$. 

\begin{mdframed}[style=thm, innerleftmargin=4pt, innerrightmargin=4pt]
\vspace{-.2cm}
\begin{theorem}
The repeated product $H_{t+1} = A \cdot  H_t$ computes an upper bound to the real influence probabilities of each infected node at step $t+1$. 
\end{theorem}
\end{mdframed}

\begin{proof}
{\footnotesize
\begin{align}
\hat{p}^{t}(u|S^t) = A_u \cdot H_t &= \sum_{v \in \mathcal{N}(u)\cap S^t} \hat{p}_v p_{vu}\geq \label{eq:hb1} \\ 
 \sum_{v \in \mathcal{N}(u) \cap S^t} p_v p_{vu}  &\geq 1- \prod_{v \in \mathcal{N}(u) \cap S^t} (1- p_v p_{vu}) = p^t( u|S^t) 
    \label{eq:high_bound}
\end{align}}

\begin{itemize}
\item  (\ref{eq:hb1}) stems from (\ref{eq:bound1}) in the manuscript:
{\footnotesize
\begin{align*}
     \hat{p}(u|S) = A_u \cdot  X = \sum_{v \in \mathcal{N}(u)  \cap S} \frac{1}{\text{deg}(u)} = 
    \sum_{v \in \mathcal{N}(u)  \cap S} p_{vu}  \\ 
    \geq 1- \prod_{v \in \mathcal{N}(u)  \cap S} (1-p_{vu}) = p(u|S).
    \label{eq:bound}
\end{align*}}

\item (\ref{eq:high_bound}) can be proved by induction similar to \cite{zhou2015upper}.
For every $p_v \leq 1$, the base case $\sum_{v \in \mathcal{X}} p_v p_{vu} \geq 1- \prod_{v\in \mathcal{X}} (1-p_v p_{vu})$ is obvious for $|\mathcal{X}| =1$. For $|\mathcal{X}| > 1$, we have:
{\footnotesize
\begin{align}
    1- \prod_{v\in \mathcal{X}} (1-p_v p_{vu}) = 1- (1-p_{x} p_{xu}) \prod_{v \in \mathcal{X} \setminus x} (1-p_v p_{vu})  \nonumber \\
     = 1 - \prod_{v\in \mathcal{X} \setminus x} (1-p_v p_{vu}) + p_{x} p_{xu} \prod_{v\in \mathcal{X} \setminus x} (1-p_v p_{vu})  \\
    \leq  \sum_{v \in \mathcal{X} \setminus x} p_v p_{vu} + p_{x}p_{xu} \prod_{v \in \mathcal{X} \setminus x} (1-p_v p_{vu}) \nonumber \\
    \leq \sum_{v\in \mathcal{X} \setminus x} p_v p_{vu} + p_{x}p_{xu} 
    = \sum_{v\in \mathcal{X}} p_v p_{vu}.
\end{align}}
\item In (\ref{eq:high_bound})  we have $p(u|v) = p_v p_{vu}$ per definition of the IC, and thus $ p(u|S) =  1- \prod_{v \in \mathcal{N}(u) \cap S} (1-p_v p_{vu})$, where $p_v=1$ for  $v \in S^1 $, which are the initial seed set that are activated deterministically. Thus, (\ref{eq:high_bound}) stands, and these probabilities are an upper bound of the real influence probabilities. Hence, the influence spread $\hat{\sigma}(S) = \sum_{ (u,v) \in  E} p_{uv}$ is an upper bound to the real $\sigma(S)$.
\end{itemize}
\end{proof}

 In reality, due to the existence of cycles, two problems arise. Firstly, if the process is repeated, the influence of the original seeds may increase again, which comes in contrast with the independent cascade model. This can be controlled by minimizing the repetitions, e.g., four repetitions cause the original seeds to be able to reinfect other nodes in a network with triangles. 
To this end, we leverage up to three neural network layers. 
Another problem due to cycles pertains to the probability of neighbors influencing each other. In this case, the product of the complementary probabilities in Eq.~(\ref{eq:bound1}) does not factorize for the non-independent neighbors.
This effect was analyzed extensively in \cite{lokhov2019scalable}, App. B, and proved that the influence probability computed by $p(u|S)$ is itself an upper bound on the real influence probability for graphs with cycles. Intuitively, the product that represents non-independent probabilities is larger than the product of independent ones. This renders the real influence probability, which is complementary to the product, smaller than what we compute. 

We can thus contend that the estimation $\hat{p}(u|S)$ provides an upper bound on the real influence probability---and we can use it to compute an upper bound to the real influence spread of a given seed set i.e., the total number of nodes influenced by the diffusion. 
Since message passing can compute inherently an approximation of influence estimation, we can parameterize it to learn a function that tightens this approximation based on supervision.
In our neural network architecture, each layer consists of a GNN with batchnorm and dropout omitted here, and starting from $H_0 = X \in \Reals^{n \times d} $ we have:
\begin{equation}
    H_{t+1} = \text{ReLU}([H_t, AH_t] W_t).
    \label{eq:glie_layer}
\end{equation}

\noindent The readout function that summarizes the graph representation based on all nodes' representations is a summation with skip connections: 
\begin{equation}
    H^G_S= \sum_{v \in V} [H^v_0,H^v_1,\dots,H^v_t].
\end{equation}
This representation captures the probability of all nodes being active throughout each layer. The output that represents the predicted influence spread is derived by:
\begin{equation}
\hat{\sigma}(S) = \text{ReLU}(H^G_S W_o).
\label{eq:sigma}
\end{equation}

\noindent 

\begin{figure}[t]
  \centering
    \includegraphics[width=.5\textwidth]{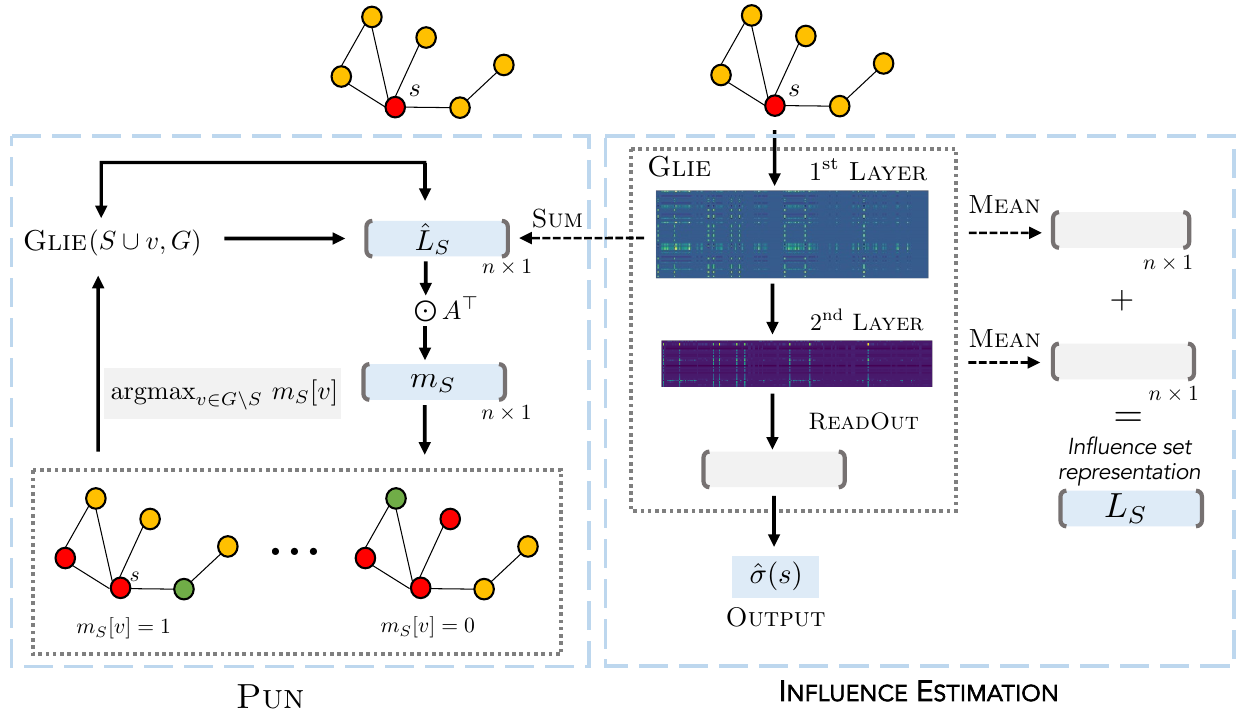}
\caption{A visual depiction of the pipeline. The layers of \textsc{Glie} are depicted by a heatmap of an actual seed during inference time, showing how the values vary through different nodes (columns).}
  \label{fig:scheme}
\end{figure}

Our loss function is a simple least squares regression.  Note that, in the case where $W_t$ is an untrained positive semidefinite Gaussian random matrix in $[0,1]$, the  representations of each layer $H_t^v$ would correspond to the upper bound of the influence probability of seed set's $t$-hop neighbors \cite{lokhov2019scalable}. This upper bound is not retained once the weights $W_t$ are trained. In our approach, the parameters of the intermediate layers $W_t$ are trained such that the upper bound is reduced and the final layer $W_o$ can combine the probabilities to derive a cumulative estimate for the total number of influenced nodes. We empirically verify this by examining the layer activations which can be seen in Fig. \ref{fig:scheme}. The heatmaps indicate a difference between columns (nodes) expected to be influenced, meaning we could potentially predict not only the number but also who will be influenced. However, since $\hat{\sigma}$ is derived by multiple layers, the relationships and thresholds to determine the exact influenced set are not straightforward. 


\subsection{{\normalfont{\textsc{Celf-Glie}}}: Cost Effective Lazy Forward with {\normalfont{\textsc{Glie}}}}
\label{sec:celf_glie}

Cost Effective Lazy Forward (\textsc{Celf}) \cite{leskovec2007cost} is an improved version of the greedy algorithm for influence maximization that exploits the property of submodularity to select seed nodes efficiently. By maintaining a sorted list of nodes based on their influence spread, CELF identifies the best node with the highest marginal gain in each iteration, resulting in significantly faster execution times compared to the greedy algorithm without sacrificing its original effectiveness.
In our case, we propose an adaptation where we substitute the original \textsc{Celf} IE based on MC IC with the output of \textsc{Glie}.
Since we do not prove the submodularity of $\hat{\sigma}$, we can not contend that the theoretical guarantee is retained, so we use this as a heuristic.
\textsc{Celf-Glie} has two main computational bottlenecks. First, it requires computing the initial IE for every node in the first step. Second, although it alleviates the need to test every node in every step, it still requires IE for at least one node in each step. Second, 
We will try to alleviate both. 

\subsection{{\normalfont{\textsc{Pun}}}: Potentially Uninfluenced Neighbors}
\label{sec:pun}
Computing the influence spread of every node in the first step is computationally demanding. We thus seek a method that can surpass this hinder and provide adequate performance.
We first utilize the activations mentioned above to define the set of influenced nodes on the step that corresponds to that layer.
Let $\hat{L}_S,L'_S \in \{0,1\}^n$ be the binary vectors with $1$s in nodes predicted to be uninfluenced and nodes predicted to be influenced respectively:
\begin{equation}
    \hat{L}_S = \mathbbm{1} \left\{ \sum_{i=0}^{d_1} H_1^i \leq 0 \right\} \hspace{.5cm}
    L'_S = \mathbbm{1} \left\{ \sum_{i=0}^{d_1} H_1^i > 0 \right\}. 
    \label{eq:infset}
\end{equation}
where $T$ is the number of layers, and $H_t^i \in \Reals^{n \times 1}$ is a column from $H_t$.
This vector contains a label for each node whose sign indicates if it is predicted to be influenced. 
$L'_S$ provides a rough estimate, but it allows for a simpler influence spread which we can optimize greedily:
\begin{equation}
    \sigma^m(S) = |L'_S|.
    \label{eq:sm}
\end{equation}
We can use $\hat{L}_S$ and message passing to predict the amount of a node's neighborhood that remains uninfluenced, i.e., the \textbf{P}otentially \textbf{U}ninfluenced \textbf{N}eighbors (\textsc{Pun}), weighted by the respective probability of influence for a node $u$, 
\begin{equation}
    m_S[u] = \sum_{v \in N(u)} A_{uv} \hat{L}_v  =  A_u^\top \cdot \hat{L}_S  \in \Reals^{n \times 1}.
    \label{eq:m}
\end{equation}
For efficiency, we can compute $m_S = A^\top \hat{L}$ which can be considered an approximation to all nodes' marginal gain on their immediate neighbors. We can thus optimize this using  $\text{arg}\max(m_S)$, as shown in Fig. (\ref{fig:scheme}). 
In order to establish that $\sigma^m$ can be optimized greedily with a theoretical guarantee of $(1-\frac{1}{e}) \text{OPT}$, in Theorem \ref{thm:submodularity}   we prove its monotonocity and submodularity.

\begin{mdframed}[style=thm, innerleftmargin=4pt, innerrightmargin=4pt]
\vspace{-.2cm}
\begin{theorem} \label{thm:submodularity}
The influence spread $\sigma^m $ is submodular and monotone.
\end{theorem}
\end{mdframed}
\begin{proof}
    The proof can be found in Appendix \ref{app:submodularity_proof}.
\end{proof}

\textsc{Pun} can be seen in the left part of Fig.~\ref{fig:scheme}. We start by setting the first seed as the node with the highest degree, which can be considered a safe assumption as in practice it is always part of seed sets. We use \textsc{Glie$(S,G)$} to retrieve $\hat{L}_S$, which we use to find the next node based on $\argmax_{v \in G \setminus S} m_{S}[v]$ and the new $\hat{L}_{S\cup \{v\}}$.
One disadvantage of \textsc{Pun} is that $\sigma^m$ is an underestimation of the predicted influence. Contrasted with the upper bound, \textsc{Dmp}, $\sigma^m$ is not as accurate as $\hat{\sigma}$, but allows us to compute efficiently a submodular proxy for the marginal gain. 
This underestimation means that a part of the network considered uninfluenced in $\hat{L}_S$ is measured as a potential gain for their neighbors, hence the ranking based on $m_S$ can be affected negatively. For this purpose, we will use adaptive full-feedback selection (AFF), where after selecting a new seed node, we remove it from the network along with nodes predicted to be influenced. It has been proved in the seminal work of \cite{golovin2011adaptive} that an AFF greedy algorithm for a submodular and monotonic function is guaranteed to have a competitive performance with the optimal policy. 
In our case, we will use an AFF update every $k$ seeds, as it adds a small computational overhead if we do it in every step. 
The benefit of \textsc{Pun} is twofold. 
Firstly, as we remove the influenced node and truncate the seed set, \textsc{Glie} produces an increasingly  more valid estimate because it performs better when the graph and seed set are smaller. 
Secondly, as the neighborhood size decreases, the effect of missed influenced nodes is diminished in $m_S$.

\section{Experimental Evaluation} \label{sec:experiments}
All the experiments are performed in a PC with an NVIDIA GPU TITAN V (12GB RAM), 256GB RAM and an Intel(R) Xeon(R) W-2145 CPU @ 3.70GHz. The source code can be found in the supplementary files.

\subsection{Influence Estimation}
\label{sec:experiments_estimation}
In order to train for the influence estimation task, we create a set of labeled samples, each consisting of the seed set $S$ and the corresponding influence spread $\sigma(S)$. We create 100 Barabasi-Albert \cite{barabasi1999emergence} and Holme-Kim \cite{holme2002growing} undirected graphs ranging from $100$ to $200$ nodes and $30$ from $300$ to $500$ nodes. 
$60\%$ are used for training, $20\%$ for validation, and $20\%$ for testing. We have used these network models because the degree distribution resembles the one of real world networks. 
The influence probabilities are assigned based on the weighted cascade model, i.e., a node $u$ has an equal probability $1/\text{deg}(u)$ to be influenced by each of her $\mathcal{N}(u)$ nodes. This model requires a directed graph, hence we turn all undirected graphs into directed ones by appending reverse edges. Though estimating influence probabilities is a problem on its own \cite{panagopoulos2020multi}, in the absence of extra data, the weighted cascade is considered more realistic than pure random assignments \cite{kempe2003maximizing}.
To label the samples, we run the \textsc{Celf} algorithm using $1,000$ Monte Carlo (MC) ICs for influence estimation, for up to $5$ seeds. Note that, we expect running $10,000$ simulations would provide a more qualitative supervision. However, on the one hand, the training time would increase exponentially, and on the other, due to the training graphs being relatively small, the difference is minuscule. The optimum seed set for size $1$ to $5$ is stored, along with $30$ random negative samples for each seed set size. This amounts to a total of $20,150$ training samples. 
Each training sample for \textsc{Glie} corresponds to a triple of a graph $G$, a seed set $S$, and a ground truth influence spread $\sigma(S)$ that serves as a label to regress on.
The random seed sets are used to capture the average influence spread expected for a seed set of about that size. This creates ``average samples" which would constitute the whole dataset in other problems. In IM however, the difference in $\sigma$ between an average seed set and the optimal can be significant, hence training solely on the random sets would render our model unable to predict larger values that correspond to the optimum. That is why we added the aforementioned samples of the optimum seed set computed using \textsc{Celf}. We deem the combination of $30$ random and $1$ optimum a more balanced form of supervision, as you expect the crucial majority of the seed sets to have an average $\sigma$.

Regarding model training, we have used a small-scale grid-search using the validation set to find the optimum batch size $64$, dropout $0.4$, number of layers $2$, hidden layer size $64$, and feature dimension $50$.
More importantly, we observed that it is beneficial to decrease the hidden layer size (by a factor of $2$) as the depth increases, i.e., go from $32$ to $16$. This means that the $1$-hop node representations are more useful compared to the $2$-hop ones and so on---validating the aforementioned conclusion that the approximation to the influence estimation in Eq. (\ref{eq:bound1}), diverges more as the message passing depth increases. The training then proceeds for 100 epochs with an early stopping of 50 and a learning rate of $0.01$. 

\begin{table}[htbp]
     \caption{Graph datasets.}
  \centering
    \resizebox{0.9\columnwidth}{!}{%
    \begin{tabular}{llcc}
    \toprule
    & \textbf{Graph}& \textbf{\# Nodes}& \textbf{\# Edges}\\
    \midrule
    \multirow{2}{.35cm}{\rotatebox{90}{\fcolorbox{green!10}{green!10}{Sim}}} & Test/Train &  $100-500$ & $950-4,810$\\
    & Large & $1,000-2,000$ & $11,066-19,076$ \\ 
    \multirow{3}{.45cm}{\rotatebox{90}{\fcolorbox{green!10}{green!10}{Small~}}} & Crime  (CR) & $829$ & $2,946$ \\ 
    & HI-II-14 (HI) & $4,165$ & $26,172$\\ 
    & GR Colab (GR) & $5,242$ & $28,980$\\ 
    \multirow{3}{.35cm}{\rotatebox{90}{\fcolorbox{green!10}{green!10}{Large}}} & Enron (EN) &  $33,697$ & $361,622$ \\ 
    & Facebook (FB) &  $63,393$ & $1,633,660$ \\ 
    & Youtube (YT) & $1,134,891$ & $5,975,246$ \\
    \bottomrule
    \end{tabular}}
     \vspace{-.4cm}
    \label{tab:small_graphs}
\end{table}

We evaluate the models in three different types of graphs. The first is the test set of the dataset mentioned above. The second is a set of $10$ power-law large graphs ($1,000$ -- $2,000$ nodes) to evaluate the capability of the model to generalize in networks that are larger by one factor. The third is a set of three real-world graphs, namely the \textit{Crime} (CR), \textit{HI-II-14} (HI), and \textit{GR collaborations} (GR). More information about the datasets is given in Table \ref{tab:small_graphs}. 

The real graphs are evaluated for varying seed set sizes, from 2 to 10, to test our model's capacity to extrapolate to larger seed set sizes. 
Due to the size of the latter two graphs (HI and GR), we take for each seed set size the top nodes based on the degree as the optimum seed set along with a $30$ random seed sets for the large simulated graphs and $3$ for the real graphs, to validate the accuracy of the model in non-significant sets of nodes. We have compared the accuracy of influence estimation with \textsc{Dmp} \cite{lokhov2019scalable}. We could not utilize the influence estimation of \textsc{Ublf} \cite{zhou2015upper} because its central condition is violated by the weighted cascade model and the computed influence is exaggerated to the point it surpasses the nodes of the network. The average error throughout all datasets and the average influence can be seen in Table \ref{tab:predict_influence}, along with the average time. 

\begin{table}[h]
\caption{Average MAE divided by the average influence and time (in seconds) throughout all seed set sizes and samples, along with the real average influence spread. }
\label{tab:predict_influence}
\centering
\resizebox{0.9\columnwidth}{!}{%
\begin{tabular}{lccccc}
\toprule
\multirow{2}{7em}{{\textbf{Graph (seeds)}}} & \multicolumn{2}{c}{\textsc{Dmp}} & & \multicolumn{2}{c}{\textsc{Glie}} 
\\
\cmidrule{2-3} \cmidrule{5-6} & \textbf{MAE} & \textbf{Time} & & \textbf{MAE} & \textbf{Time} \\
\midrule
Test ($1$ -- $5$) & $0.076$ & $0.05$ & & $\textbf{0.046}$ & $\textbf{0.0042}$ \\ 
Large ($1$ -- $5$) & $\textbf{0.086}$ & $0.44$ & &$0.102$ & $\textbf{0.0034}$ \\ 
CR ($1$ -- $10$) &  $\textbf{0.009}$  & $0.11$ & & $0.044$ & $\textbf{0.0029}$ \\ 
HI ($1$ -- $10$) & $\textbf{0.041}$ & $2.84$ & & $0.056$ & $\textbf{0.0034}$	\\
GR ($1$ -- $10$) & $0.122$ & $4.32$ & & $\textbf{0.084}$ & $\textbf{0.0042}$ \\ 
\bottomrule
\end{tabular}}

\vspace{.8cm}

\caption{IM for 20 seeds with \textsc{CELF}, using the proposed (\textsc{Glie}) substitute for influence estimation and evaluating with $10,000$ MC independent cascades (IC). }
\label{tab:small_im}
\centering
\resizebox{0.9\columnwidth}{!}{%
\begin{tabular}{@{\extracolsep{4pt}}lccccc}
\toprule
\multirow{2}{4em}{{\textbf{Graph (seeds)}}} & 
\multirow{2}{3.4em}{{\centering \textbf{Seed \\ overlap}}} & 
\multicolumn{2}{c}{\textsc{Dmp-Celf}} & \multicolumn{2}{c}{\textsc{Glie-Celf}} \\
\cmidrule{3-4} \cmidrule{5-6}
 &  & \textbf{Infl} & \textbf{Time} & \textbf{Infl} & \textbf{Time} \\
\midrule
CR~($20$) & $14$ & $221$ & $83$ & $\textbf{229}$ & $\textbf{1.0}$ \\ 
HI~($20$) & $13$ & $1,235$ & $8,362$ & $\textbf{1,281}$ & $\textbf{5.49}$ \\ 
GR~($20$) & $12$ & $295$ & $16,533$ &  $\textbf{393}$ & $\textbf{7.01}$ \\ 
\bottomrule
\end{tabular}}
\end{table}

We evaluate the retrieved seed set using the independent cascade, and the results are shown in Table \ref{tab:small_im}. We should underline here that this task would require more then $3$ hours for the \textit{Crime} dataset and days for \textit{GR} using the traditional approach with $1,000$ MC IC. As we can see in Table \ref{tab:small_im}, \textsc{Glie-Celf} allows for a significant acceleration in computational time, while the retrieved seeds are more effective.
Moreover, in \textsc{Celf}, the majority of time is consumed in the initial computation of the influence spread, i.e., the overhead to compute 100 instead of the $20$ seeds shown in Table \ref{tab:small_im}, amounts to $0.11$, $0.22$ and $0.19$ seconds for the three datasets respectively.

\subsection{Influence Maximization}
\label{sec:experiments_maximization}

\begin{table*}[h!]
 \caption{Influence spread computed by 10,000 MC ICs for $200$ seeds.}
 \label{tab:seeds200}
 \centering
 \resizebox{0.9\textwidth}{!}{%
 \begin{tabular}{lcc||cccccc}
 \toprule
\textbf{Graph} & \textsc{Glie-Celf} &\textsc{Pun} & \textsc{K-core}& \textsc{Pmia}& \textsc{DegDisc} &
 \textsc{Imm} & \textsc{DeepIS-Celf}& \textsc{FINDER}\\
 \midrule
CR & $\bf{661}$  & \underline{$657$} & $647$ &$656$ & $644$ &  $650$& $501.61$ & $642$\\ 
GR & \underline{$1,617$} &$\bf{1,626}$  & $701$ &$1,566$ & $1415$ & $835.40$ & \underline{$1,617$} &$1,286$\\ 
HI &\underline{$2,685$} & $\bf{2,688}$  & $2,540$ & $2,685$ & $2,614$ &  $2,668$ & $1602.5$& $2,625$\\ 
EN& \underline{$17,601$}   & $\bf{17,614}$ & $13,015$ & $17,534$ & $16,500$ & $17,497$ &-& $17,244$ \\ 
FB & \underline{$10,981$}    & $10,626$ & $6,434$ & $7,688$ & $10,309$ & $\bf{11,007}$ &-& $10,801$\\
YT &  $\underline{246,439}$  & $244,579$& $110,409$ & $242,057$& $236,726$  & \bf{$247,178$}&-   &$50,435$\\
   \bottomrule
  \end{tabular}
  }
  \end{table*}

 \begin{table}
   \caption{Computational time in seconds (vs Algorithms).}
   \label{tab:comp_time}
 \centering
 \begin{tabular}{lcccc}
 \toprule
 \textbf{Graph} &  \textsc{Glie-Celf}  & \textsc{Pun} & \textsc{Imm} & \textsc{FINDER}\\
 \midrule
  CR& $2.00$  &  \underline{$0.25$}& $\bf{0.19}$ &$0.41$\\  
  GR &$4.55$ & $\bf{0.26}$ & $\underline{0.95}$&$2.36$ \\  
  HI  & $2.19$ & $\bf{0.27}$ & $1.29$ & $1.01$ \\    
  EN& $15.49$  & $\bf{0.97}$ & $10.47$ & $9.30$\\  
  FB & $287.7$  & $\bf{3.1}$ & $171.25$ & \underline{$56.80$} \\  
  YT &$151.33$ & $\bf{28.92}$ &\underline{$82.13$} &$191.00$\\  
  \bottomrule
  \end{tabular}
\end{table}

\begin{table}[t!]
\caption{Computational time in seconds (vs heuristics).}
 \label{tab:heuristics_time}
\centering
 \begin{tabular}{lcccc}
 \toprule
 \textbf{Graph} & \textsc{Pun} & \textsc{DegDisc} &\textsc{K-core} & \textsc{PMIA}\\
 \midrule
CR &
$0.25$ &$0.21$ &\underline{$0.06$} & $\textbf{0.04}$ \\ 
GR 
&\underline{$0.26$} &$0.80$ &$\textbf{0.13}$ &$1.5$  \\ 
HI
&0.27 &1.36 &\underline{$0.14$} & $\textbf{0.12}$  \\ 
EN 
& $\textbf{0.97}$ &$26.74$ & \underline{$2.06$} &$2.17$ \\ 
FB  & 
$\textbf{3.1}$  & $22.77$ & \underline{$9.29$} & $10.62$ \\ 
YT 
& $\textbf{28.92}$ & $4006.29$ & \underline{$54.38$} & $74.91$ \\ 
  \bottomrule
  \end{tabular}
 \end{table}

Our main benchmark is a state-of-the-art IM method, \textsc{Imm} \cite{tang2015influence}, which capitalizes on reverse reachable sets \cite{borgs2014maximizing} to estimate influence. Specifically, it produces a series of such sketches and uses them to approximate the influence spread without simulations. This results in remarkable acceleration with a theoretical guarantee. Note that, \textsc{Imm} is considered one of the state-of-the-art algorithms and surpasses various heuristics \cite{jung2012irie}. We set $e =0.5$ as proposed by the authors. We also compare with \textsc{Finder}, which is analyzed in Section \ref{sec:related_work}, and with the most well-known heuristic methods for the Independent Cascade. \textsc{Pmia} \cite{wang2012scalable} computes the influence spread based on local approximations. \textsc{DegreeDiscount} \cite{chen2009efficient} builds a seed set using the node's degree, which is recomputed based on the current seed set and its influence. Finally, \textsc{K-cores} \cite{malliaroscore} is the a graph degeneracy metric that uncovers nodes that are part of densely connected subgraphs.

The results for the influence spread of $200$ seeds as computed by simulations of MC ICs can be seen in Table \ref{tab:seeds200}, while the time results are shown in Tables \ref{tab:comp_time} and \ref{tab:heuristics_time}. The best result is in bold and the second best is underlined. 
One can see that \textsc{Glie-Celf} exhibits overall superior influence quality compared to the rest of the methods, but is quite slower.  
\textsc{Pun} requires only one influence estimation in every step and no initial computation. It exhibits 3 to 60 times acceleration compared to \textsc{Imm} while its computational overhead moving from smaller to larger graphs is sublinear to the number of nodes. In terms of influence quality, \textsc{Pun} is first or second in the majority of the datasets and this effect becomes more clear as the seed set size increases. \textsc{DegDisc} is faster than \textsc{Pun} in smaller graphs but slower in larger and overall worse in seed set quality. \textsc{Pmia} provides medium seed set quality but is computationally inefficient. 
\textsc{Imm} is clearly not the fastest method, but it is very accurate, especially for smaller seed set sizes. 
\textsc{Finder} exhibits the least accurate performance, which is understandable given that it solves a relevant problem and not exactly IM  for IC. The computational time presented is the time required to solve the node percolation, in which case it may retrieve a bigger seed set than 100 nodes. Thus, we can hypothesize it is quite faster for a limited seed set, but the quality of the retrieved seeds is the least accurate among all methods.
Overall, we can contend  that \textsc{Pun} provides the best accuracy-efficiency tradeoff from the examined methods.

\textsc{DeepIS}, as analyzed in related work, resembles \textsc{Glie}, in that it computes influence estimation using a neural network. We follow the authors' methodology and train the model using their code on the proposed Cora ML \cite{xia2021deepis}. We use it as an influence estimation oracle in \textsc{Celf}, similar to \textsc{Glie-Celf}. Unfortunately, it is infeasible to scale in the larger datasets due to the need for explicit powers of the influence matrices that required more than 24 GB of GPU RAM. 
For comparison, this model required  $501.61$, $835.4$, and $1,602.5$ seconds for the CR, GR, and HI datasets, respectively. This further supports the superiority of \textsc{Pun}.

\begin{table}[t!]
\caption{\textsc{Pun} CPU vs. GPU time (sec).}
\label{tab:cpu_time}
\begin{minipage}{.5\textwidth}
\centering
\resizebox{0.75\columnwidth}{!}{%
\begin{tabular}{cccccc}
\toprule
\textbf{Graph} & \textsc{Pun} GPU & ~& \textsc{Pun} CPU &~& IMM\\
\midrule 
CR  & 0.15 && 0.17 && 0.13\\ 
GR  & 0.17 && 0.27 && 0.57\\ 
HT  & 0.17 && 0.20 && 0.56\\ 
EN  & 0.52 && 2.44 && 4.78\\  
FB  & 1.42 && 17.5 && 69.9 \\ 
YT  & 13.2 && 97.5 && 55.4\\ 
\bottomrule
\end{tabular}}
\end{minipage}
\end{table}

\pgfplotsset{every tick label/.append style={font=\LARGE}}
\pgfplotsset{every axis/.append style={font=\LARGE}}

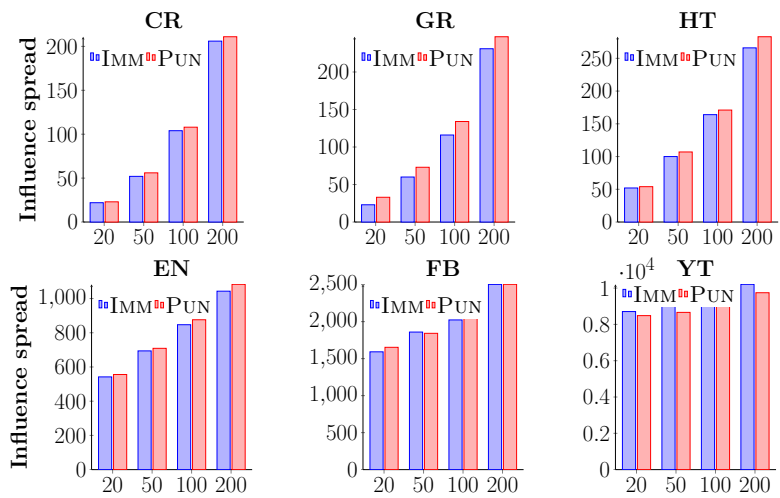
\begin{figure}[t]
\centering
\begin{tabular}{cc}
\begin{tikzpicture}[scale=.42]
\begin{axis}  
[  ybar,
title={CR},
    bar width=0.5cm, 
    x=1.5cm,
    enlarge x limits={abs=0.8cm},
    legend style={draw=none},
    legend style={at={(0.3,0.95)},  
      anchor=north,legend columns=-1},     
    ylabel={{Influence spread}},
    y label style={at={(-0.1,0.5)}},
    symbolic x coords={20, 50, 100, 200},  
    xtick=data,
    axis y line=left,
    nodes near coords align={vertical},
    axis x line*=bottom,
    ymin=0
]  
\addplot 
	coordinates {(20, 22) (50, 52) (100, 104) (200,206)};
\addplot
	coordinates {(20, 23) (50, 56) (100, 108) (200,211)};
\legend{\textsc{Imm},\textsc{Pun}}
\end{axis}
\end{tikzpicture}
&
\begin{tikzpicture}[scale=.42]
\begin{axis}  
[  ybar,
    title={GR},
    bar width=0.5cm, 
    x=1.5cm,
    enlarge x limits={abs=0.8cm},
    legend style={draw=none},
    legend style={at={(0.3,0.95)},  
      anchor=north,legend columns=-1},     
    ylabel={Influence spread},
    y label style={at={(-0.1,0.5)}},
    symbolic x coords={20, 50, 100, 200},  
    xtick=data,
    axis y line=left,
    nodes near coords align={vertical},
    axis x line*=bottom,
    ymin=0
]  
\addplot 
	coordinates {(20, 23) (50, 60) (100, 116) (200,231)};
\addplot
	coordinates {(20, 33) (50, 73) (100, 134) (200,247)};
\legend{\textsc{Imm},\textsc{Pun}}
\end{axis}
\end{tikzpicture}
\\

\begin{tikzpicture}[scale=.42]
\begin{axis}  
[  ybar,
title={HT},
    bar width=0.5cm, 
    x=1.5cm,
    enlarge x limits={abs=0.8cm},
    legend style={draw=none},
    legend style={at={(0.3,0.95)},  
      anchor=north,legend columns=-1},     
    ylabel={Influence spread},
    y label style={at={(-0.1,0.5)}},
    symbolic x coords={20, 50, 100, 200},  
    xtick=data,
    axis y line=left,
    nodes near coords align={vertical},
    axis x line*=bottom,
    ymin=0
]  
\addplot 
	coordinates {(20, 52) (50, 100) (100, 164) (200,266)};
\addplot
	coordinates {(20, 54) (50, 107) (100, 171) (200,283)};
\legend{\textsc{Imm},\textsc{Pun}}
\end{axis}
\end{tikzpicture}
&
\begin{tikzpicture}[scale=.42]
\begin{axis}  
[  ybar,
    title={EN},
    bar width=0.5cm, 
    x=1.5cm,
    enlarge x limits={abs=0.8cm},
    legend style={draw=none},
    legend style={at={(0.3,0.95)},  
      anchor=north,legend columns=-1},     
    ylabel={Influence spread},
    y label style={at={(-0.1,0.5)}},
    symbolic x coords={20, 50, 100, 200, 250},  
    xtick=data,
    axis y line=left,
    nodes near coords align={vertical},
    axis x line*=bottom,
    ymin=0
]  
\addplot 
	coordinates {(20, 542) (50, 694) (100, 847) (200,1043)};
\addplot
	coordinates {(20, 556) (50, 709) (100, 876) (200,1083)};
\legend{\textsc{Imm},\textsc{Pun}}
\end{axis}
\end{tikzpicture}
\\

\begin{tikzpicture}[scale=.42]
\begin{axis}  
[  ybar,
title={FB},
    bar width=0.5cm, 
    x=1.5cm,
    enlarge x limits={abs=0.8cm},
    legend style={draw=none},
    legend style={at={(0.3,0.95)},  
      anchor=north,legend columns=-1},     
    ylabel={Influence spread},
    y label style={at={(-0.1,0.5)}},
    symbolic x coords={20, 50, 100, 200},  
    xtick=data,
    axis y line=left,
    nodes near coords align={vertical},
    axis x line*=bottom,
    ymin=0
]  
\addplot 
	coordinates {(20, 1592) (50, 1859) (100, 2023) (200,2503)};
\addplot
	coordinates {(20, 1653) (50, 1842) (100, 2114) (200,2504)};
\legend{\textsc{Imm},\textsc{Pun}}
\end{axis}
\end{tikzpicture}
&
\begin{tikzpicture}[scale=.42]
\begin{axis}  
[  ybar,
    title={YT},
    bar width=0.5cm, 
    x=1.5cm,
    enlarge x limits={abs=0.8cm},
    legend style={draw=none},
    legend style={at={(0.3,1.02)},  
      anchor=north,legend columns=-1},     
    ylabel={Influence spread},
    y label style={at={(-0.1,0.5)}},
    symbolic x coords={20, 50, 100, 200, 250},  
    xtick=data,
    axis y line=left,
    nodes near coords align={vertical},
    axis x line*=bottom,
    ymin=0
]  
\addplot 
	coordinates {(20, 8708) (50, 9012) (100, 9495) (200,10207)};
\addplot
	coordinates {(20, 8484) (50, 8662) (100, 9077) (200,9745)};
\legend{\textsc{Imm},\textsc{Pun}}
\end{axis}
\end{tikzpicture}
\end{tabular}
\caption{\textsc{Pun} vs. \textsc{Imm} for IC with $p=0.01$.}
\label{fig:imm_pun_ic}
\end{figure}

Furthermore, we compare \textsc{Imm} and \textsc{Pun} on the same graphs with uniform influence probabilities $p=0.01$ in Figure \ref{fig:imm_pun_ic}, as a substitute to the weighted cascade assignment. We observe that \textsc{Pun} outperforms \textsc{Imm}.
Finally, we performed an experiment to compare \textsc{Pun} without the use of GPU for 100 seeds. The results are reported in Table \ref{tab:cpu_time}. It is visible that GPU provides a substantial acceleration, but \textsc{Pun} remains the faster option even without it.
 
\section{Conclusion}
\label{sec:conclusion}
We have proposed \textsc{Glie}, a GNN-based model for influence estimation. We showcase its accuracy in that task and further utilize it to address the problem of IM. We developed two methods based on the representations and the predictions of \textsc{Glie}. \textsc{Glie-Celf}, an adaptation of a classical algorithm that surpasses SOTA but with significant computational overhead, and textsc{Pun}, a submodular function that acts as proxy for the marginal gain and can be optimized adaptively, striking a balance between efficiency and accuracy. 

For a typical IM algorithm, it is not straightforward to consider the topic of the information shared or the user's characteristics \cite{chen2016real}. A significant practical advantage of a neural network approach is the easy incorporation of such complementary data by adding the corresponding embeddings in the input, as has been done in similar settings \cite{tian2020deep}. We thus deem an experiment with contextual information a natural next step, given a proper dataset.  
Finally, we also plan to examine the potential of training a reinforcement learning module, i.e., receiving real feedback from each step of the diffusion that could update both the \textsc{Q-Net} and \textsc{Glie}. This would allow the model to adjust its decisions based on the partial feedback received during the diffusion.

\appendices
\section{Proof of Theorem \ref{thm:submodularity}}

\label{app:submodularity_proof}
For the purposes of the proof, 
$P\in \{1\} ^{hd \times 1}$ and we define the support function $\mathcal{S}(v) = \{i \in [1,n], v_i \neq 0  \}$ 
as the set of indices of non zero rows in a matrix such as $X_i$, of layer $i$. Let $R$ represent ReLU and $b_{tr}, st_{tr}$ the mean and standard deviation computed by the batchnorm. 
\vspace{-1cm}
\begin{proof}
Monotonocity, $ \forall i<j, S_i \subset S_j$:
{\footnotesize 
\begin{align}
\mathcal{S}( X_j )   \supset  \mathcal{S}( X_i ) &\Rightarrow \mathcal{S}(  X_j W )   \supseteq \mathcal{S}( X_i W ) 
\label{eq:sb2}\\
 \mathcal{S}( A  X_j W )  \supseteq \mathcal{S}( A X_i W ) &\Rightarrow  \mathcal{S}( (R(A  X_j W ) - b_{tr} )/st_{tr} )   \nonumber\\  \supseteq \mathcal{S} (( R(A  X_i W) - b_{tr})/st_{tr} ) \label{eq:sb4}\\
 \mathcal{S}  ( H_j ) \supseteq \mathcal{s}  ( H(S_i) ) & 
 \Rightarrow \mathcal{S}  ( H_j P ) \supseteq \mathcal{S}  ( H_i P )  \label{eq:sb7}\\ 
| \mathbbm{1}_{> 0} \left\{  H_j P   \right\} |  \geq | \mathbbm{1}_{> 0} \left\{ H_i P \right\} | &\Rightarrow \sigma^m(S_j)  \geq \sigma^m(S_i)\label{eq:sb10}
\end{align}}%


(\ref{eq:sb2}) First subset is by definition. Second is because $X_j$ is a convex hull that contains $X_i$ \cite{boyd2004convex}. We multiply both sides by a real matrix $W 
\in \Reals^{d\times hd}$ which can equally dilate both convex hulls in terms of direction and norm. This transformation cannot change the sign of the difference between the elements of $X_i$ and $X_j$
and hence cannot interfere with the support of $X_j$ over $X_i$. This becomes more obvious for $X \in \{0,1\}^{n \times 1}$ and $W 
\in \Reals^{1 \times 1}$.  Note that both can result in zero matrices so we use subset or equal.
(\ref{eq:sb4}) $A$ is a non-negative matrix and ReLU is a nonnegative monotonically increasing function. Subtract by the same number and divide by the same positive number in the right inequality.
(\ref{eq:sb7}) Definition in Eq. (\ref{eq:glie_layer}); $P$ is positive.
(\ref{eq:sb10}) By definition of the support and of $L'_S$.
\end{proof}

For the proof of submodularity we have to define $X_{iu} = X_{S_i \cup u}, u \in V$ and note by the definition of the input that $ |X_{ju} - X_j| = |X_{iu} - X_i| $ for the $l_1$ norm (sum of all elements):

\begin{proof}
Submodularity $ \forall i<j, S_i \subset S_j$:
{\scriptsize
\begin{align}
&|X_{ju} - X_j|  = |X_{iu} - X_i| \Rightarrow 
 |A (X_{ju} - X_j) |  =  |A (X_{iu} - X_i) | \label{eq:sb13}\\
 &| AX_{ju}W - AX_jW |  =  |A X_{iu}W - AX_iW | \label{eq:sb15}\\
 &R(| AX_{ju}W - AX_jW |) - 2b_{tr}  = R(|A X_{iu}W - AX_iW |-2b_{tr}) \label{eq:sb16}\\
 &| R(AX_{ju}W )- R(AX_jW) - 2b_{tr}| = \nonumber\\ 
 &| R(AX_{iu}W )- R(AX_iW) - 2b_{tr}|  \label{eq:sb17}\\
 &\mathcal{S}( R(AX_{ju}W )- R(AX_jW) - 2b_{tr}) = \nonumber \\&\mathcal{S}( R(AX_{iu}W )- R(AX_iW) - 2b_{tr}) \label{eq:sb18}\\
 &\mathcal{S}( R(AX_{ju}W  -b_{tr}))- \mathcal{S}( R(AX_jW) -b_{tr})  \subseteq  \nonumber \\ &\mathcal{S}( R(AX_{iu}W -b_{tr} ))- \mathcal{S}(R(AX_iW) - b_{tr}) \label{eq:sb19}\\
 &\mathcal{S}(H_{ju})-\mathcal{S}(H_{j})  \subseteq  \mathcal{S}(H_{iu})-\mathcal{S}(H_{i}) \label{eq:sb20}\\
 &\sigma^m(S^j \cup \{u\}) - \sigma^m(S^j) \leq \sigma^m(S^i \cup \{u\}) - \sigma^m(S^i)  & \label{eq:sb21}
\end{align}}%

(\ref{eq:sb15}) Distributive property and $W$ similar to multiply by $A$.
\item (\ref{eq:sb19}) The norm of the difference is distributed equally, but the right hand difference has as least the same or more positive elements because the norm of $A$, which is stochastic, is bounded by $V$ hence $X_u$ can give up to the same gain to $AX_j$ and $AX_i$, the same number $b_{tr}$ is subtracted, and more elements are activated by $X_j$ then $X_i$ as shown in (\ref{eq:sb10}). (\ref{eq:sb20}) We skipped dividing by $st_{tr}$ for brevity.(\ref{eq:sb21}) Similar steps as (\ref{eq:sb7}) -  (\ref{eq:sb10}).
\end{proof}

Regarding the approximation of the marginal gain we first show that the node corresponding to the maximum $m_S$ gives the maximum $L'_{ju}$:
$A'_u\hat{L}_j \geq A'_v\ \hat{L}_i  \Rightarrow	 L'_{ju} \geq L'_{iv}$.
{\footnotesize
\begin{equation*}
    A'_u\hat{L}_j = \sum_{v \in N(u)} A'_{uv} \hat{L}_j[v] = \sum_{v \in N(u)} A_{uv} L'_j[u] = \sum_{v \in N(u)} A_{uv} X_{ju}.
\end{equation*}}
This means that $m_S$ gives the node $u$ that improves the biggest number of rows in $AX_{ju}$ that are not already considered influenced. Since we know from Eq.~(\ref{eq:sb10}) that  $AX_{iu} \geq AX_{iv} \Rightarrow |L'_{iu}| \geq |L'_{iv}|$, the claim concludes. Hence, choosing the best node using the marginal gain approximation is as good as the real influence spread. Now we prove the submodularity of the proposed marginal gain.

\begin{proof}
Submodularity for the approximation of the marginal gain, $ \forall i<j, S_i \subset S_j$, starting from (\ref{eq:sb10}):
{\footnotesize
\begin{align}
| \mathbbm{1}_{> 0} \left\{  H_j P   \right\} | &\geq | \mathbbm{1}_{> 0} \left\{ H_i P \right\} |  \nonumber \\
|\mathbbm{1}_{\leq 0} \left\{ H_j P  \right\} |  &\leq | \mathbbm{1}_{\leq 0} \left\{ H_i P  \right\} |  \label{eq:sb22}\\ 
A'_u\hat{L}_j \leq  A'_u\hat{L}_i  &\rightarrow 
m_{S_j}[u]  \leq  m_{S_i}[u]    \label{eq:sb24}\\
(|L'_j| +m_{S_j}[u]) - |L'_j| & \leq (|L'_i|+ m_{S_i}[u]) - |L'_i| \label{eq:sb25}\\
\sigma^m(S^j \cup \{u\}) - \sigma^m(S^j)  & \leq \sigma^m(S^i \cup \{u\}) - \sigma^m(S^i) \label{eq:sb26}  
\end{align}}
(\ref{eq:sb22}) Complementarity between elements that are $ \leq 0$ and $ >0$.
(\ref{eq:sb24}) Definition in Eq. (\ref{eq:infset}) and multiply with non-negative row $u$ from matrix $A'$,  Definition in Eq. (\ref{eq:m}).
(\ref{eq:sb26}) By definition of $\sigma^m$ in Eq. (\ref{eq:sm}) and the marginal gain of $u$.
\end{proof}

\bibliographystyle{IEEEtran}
\bibliography{glie}

\end{document}